\RequirePackage{fix-cm}
\documentclass{svjour3}                     
\smartqed  
\usepackage{flafter}%
\usepackage{amsmath}
\usepackage{graphicx}
\usepackage{epstopdf}
\usepackage[utf8]{inputenc}
\usepackage{mathptmx}      
\usepackage[colorlinks=true, pdfborder={0 0 0}]{hyperref}
\usepackage{latexsym}

\begin{document}
\title{Using Artificial Neural Network Techniques for Prediction of Electric Energy Consumption
}


\author{Hasan M. H. Owda \and Babatunji Omoniwa \\ \and Ahmad R. Shahid  \and Sheikh Ziauddin
}


\institute{ H. M. H. Owda \and B. Omoniwa \and  A. R. Shahid \and S. Ziauddin \at
              COMSATS Institute of Information Technology, Park Road, Islamabad, Pakistan \\
              Tel.: +92-33-65879709\\
              \email{hasanouda@gmail.com}           
           \and
           B. Omoniwa\at
           National Mathematical Centre, Abuja, Nigeria \\
            \email{tunjiomoniwa@gmail.com}
            \and
           A. R. Shahid\at
            \email{ahmadrshahid@comsats.edu.pk}
            \and
           S. Ziauddin\at
            \email{sheikh.ziauddin@comsats.edu.pk}
}

\date{Received: date / Accepted: date}

\maketitle

\begin{abstract}
Due to imprecision and uncertainties in predicting real world problems, artificial neural network (ANN) techniques have become increasingly useful for modeling and optimization. This paper presents an artificial neural network approach for forecasting electric energy consumption. For effective planning and operation of power systems, optimal forecasting tools are needed for energy operators to maximize profit and also to provide maximum satisfaction to energy consumers. Monthly data for electric energy consumed in the Gaza strip was collected from year 1994 to 2013. Data was trained and the proposed model was validated using 2-Fold and K-Fold cross validation techniques. The model has been tested with actual energy consumption data and yields satisfactory performance.
\end{abstract}

\section{Introduction}
\label{intro}
Numerous techniques for forecasting electric energy consumption have been proposed in the last few decades. For operators, energy consumption (load) forecast is useful in effectively managing power systems. Consumers can also benefit from the forecasted information in order to yield maximum satisfaction. In addition to these economic reasons, load forecasting has also been used for system security purposes. When deployed to handle system security problems, it provides expedient information for detecting vulnerabilities in advance.

Forecasting energy consumed within a particular geographical area greatly depends on several factors, such as, historical load, mean atmospheric temperature, mean relative humidity, Population, GDP per Capita. Over the years, there has been rapid growth annually of about 10\% from year 1999 to 2005 for energy demand in the Gaza strip. With about 75\% of energy demands from service and household sectors, these demands are barely met \cite{RefJ18,RefJ19}. In order to meet these demands and efficiently utilize the limited energy, it is imperative to observe historic trends and make futuristic plans based on past data.

In the past, computationally easier approaches like regression and interpolation, have been used, however, this methods may not give sufficiently accurate results. As advances in technology and sophisticated tools are made, complex algorithmic approaches are introduced and more accuracy at the expense of heavy computational burden can be observed. Several algorithms have been proposed by several researchers to tackle electric energy consumption forecasting problem. Previous works can be grouped into three\cite{RefJ1}: \begin{description}
                         \item[\textit{\textbf{Time Series Approach:}}] In this approach, the trend for electric energy consumption is handled as a time series signal. Future consumption is usually predicted based on various time series analysis techniques. However, time series approach is characterized with prediction inaccuracies of prediction and numerical instability. This inaccurate results is due to the fact the approach does not utilize weather information. Studies have shown that there is a strong correlation between the behavior of energy consumed and weather variables. Zhou R. \textit{et al}. \cite{RefJ9} proposed a data driven modeling method using time series analysis to predict energy consumed within a building. The model in \cite{RefJ9} was applied on two commercial building and is limited to energy prediction within a building. Basu K. \textit{et al}. \cite{RefJ12} also used the time series approach to predict appliance usage in a building for just an hour.

                             Simmhan Y. \textit{et al}. \cite{RefJ10} used an incremental time series clustering approach to predict energy consumption. This method in \cite{RefJ10} was able to minimize the prediction error, however, very large number of data points were required. Autoregressive integrated moving average (ARIMA) is a vastly used time series approach. ARIMA model was used by Chen J. \textit{et al}. \cite{RefJ11} to predict energy consumption in Jiangsu province in China based on data collected from year 1985 to 2007. The model \cite{RefJ11} was able to accurately predict the energy consumption, however it was limited to that environment. The Previous works on time series usually use computationally complex matrix-oriented adaptive algorithms which, in most scenarios, may become unstable.
                         \item[\textit{\textbf{Functional Based Approach:}}] Here, a functional relationship between a load dependent variable (usually weather) and the system load is modelled. Future load is then predicted by inserting the predicted weather information into the pre-defined functional relationship. Most regression methods use functional relationships between weather variables and up-to-date load demands. Linear representations are used as forecasting functions in conventional regression methods and this method finds an appropriate functional relationship between selected weather variables and load demand. Liu D. \textit{et al}. \cite{RefJ13} proposed a support vector regression with radial basis function to predict energy consumption in a building. The approach in \cite{RefJ13} was only able to forecast the energy consumed due to lighting for some few hours.

                             In \cite{RefJ14}, a grey model, multiple regression model and a hybrid of both were used to forecast energy consumption in Zhejiang province of China. Yi W. \textit{et al}. \cite{RefJ17} proposed an LS-SVM regression model to also forecast energy consumption. However, these models were limited to a specific geographic area.
                         \item[\textit{\textbf{Soft Computing Based Approach:}}] This is a more intelligent approach that is extensively being used for demand side management. It includes techniques such as fuzzy logic, genetic algorithm and artificial neural networks (ANN) \cite{RefJ15,RefJ16,RefJ4,RefJ2,RefJ5}. The ANN approach is based on examining the relationship that exist between input and output variables. ANN approach was used in \cite{RefJ2} to forecast regional load in Taiwan. Empirical data was used to effectively develop an ANN model which was able to predict the regional peak load. Catalão J. P. S. \textit{et al}. \cite{RefJ5} used the ANN approach to forecast short-term electricity prices. Levenberg-Marquardt's algorithm was used to train data and the resulting model \cite{RefJ5} was able to accurately forecast electricity prices. However, it was only able to predict electricity prices for about 168 hours.

                             Pinto T. \textit{et al}. \cite{RefJ6} also worked on developing an ANN model to forecast electricity market prices with a special feature of dynamism. This model \cite{RefJ6} performs well when a small set of data is trained, however, it is likely to perform poorly with large number of data due to the computational complexities involved. Load data from year 2006 to 2009 were gathered and used to develop an ANN model for short-term load forecast in \cite{RefJ7}. In \cite{RefJ8}, ANN Hybrid with Invasive Weed Optimization (IWO) was employed to forecast the electricity prices in the Australian Market. The hybrid model \cite{RefJ8} showed good performance, however, the focus was on predicting electricity prices in Australia. Most of the ANN models developed in existing work considered some specific geographic area\cite{RefJ2,RefJ8}, some models were able to forecast energy consumption for buildings\cite{RefJ13,RefJ14,RefJ15} and only for few hours\cite{RefJ12,RefJ5}.
                       \end{description}
This study used available historical data\footnote{Data was collected from the Palestinian Bureau of Statistic \cite{RefJ18} and the World bank website \cite{RefJ19}} from year 1994 to 2013 (but trained data from year 1994 to 2011) in determining a suitable model. The resulting model from training will be used to predict electric energy consumption for future years\footnote{Predictions for 2012 and 2013 were compared with actual data of corresponding years in Section \ref{sec:4}}, while the error criteria such as mean squared error (MSE), root mean squared error (RMSE), mean absolute error (MAE) and mean absolute percentage error (MAPE) are used as measures to justify the appropriate model \cite{RefJ5,RefJ7,RefJ3}. The model was used to predict the behavior for year 2012 and 2013. The remainder of this paper is organized as follows: Section \ref{sec:2} gives a brief description of the ANN concept. Section \ref{sec:3} presents the ANN approach used to analyse our data. Section \ref{sec:4} evaluates the performance of the ANN model. Section \ref{sec:5} draws conclusions.
\section{ANN Concept}
\label{sec:2}
ANN is a system based on the working principles of biological neural networks, and is defined as a mimicry of biological neural systems. ANN's are at the vanguard of computational systems designed to create or mimic intelligent behavior. Unlike classical Artificial Intelligence (AI) systems that are aimed at directly emulating rational and logical reasoning, ANN’s are targeted at reproducing the causal processing mechanisms that give results to intelligence as a developing property of complex systems.

ANN systems have been designed for fields such as capacity planning, business intelligence, pattern recognition, robotics. .  In computer science and engineering, ANN techniques has gained a lot of grounds and is vastly deployed in areas such as forecasting, data analytics and even data mining. The science of raw data examination with the aim of deriving useful conclusions can simply be defined as data analytics. On the other hand, data mining describes the process of determining new patterns from large data sets, by applying a vast set of approaches from statistics, artificial intelligence, or database management. Forecasting is useful in predicting future trends with reliance on past data. The focus on this paper will be on using the ANN approach in forecasting energy consumption. Practically, ANN provides accurate approximation of both linear and non-linear functions.
\begin{figure*}[h]
 \begin{center}
 \includegraphics[width=1\textwidth]{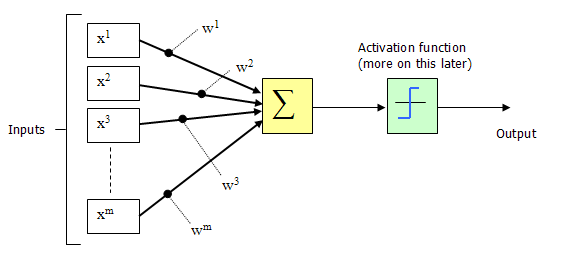} 
 \end{center}
\caption{Mathematical Abstraction of the ANN}
\label{fig:1}       
\end{figure*}

A mathematical abstraction of the internal structure of a neuron is shown in Fig. \ref{fig:1}. Despite the presence of noise or incomplete information, it is possible for neurons to learn the behaviour or trends and consequently, make useful conclusions. Usually, a neural network is trained to perform a specific function by adjusting weight values between elements as seen in Fig. \ref{fig:2}(a). The neural network function is mainly determined by the connections between the elements \cite{RefJ7}. By observing the data, it is possible for the ANN to make accurate predictions. Unlike other forecasting approach, the ANN technique has the ability to predict future trends with theoretically poor, but rich data set.
\section{Model Process}
\label{sec:3}
\begin{figure*}[h]
 \begin{center}
 \includegraphics[width=1\textwidth]{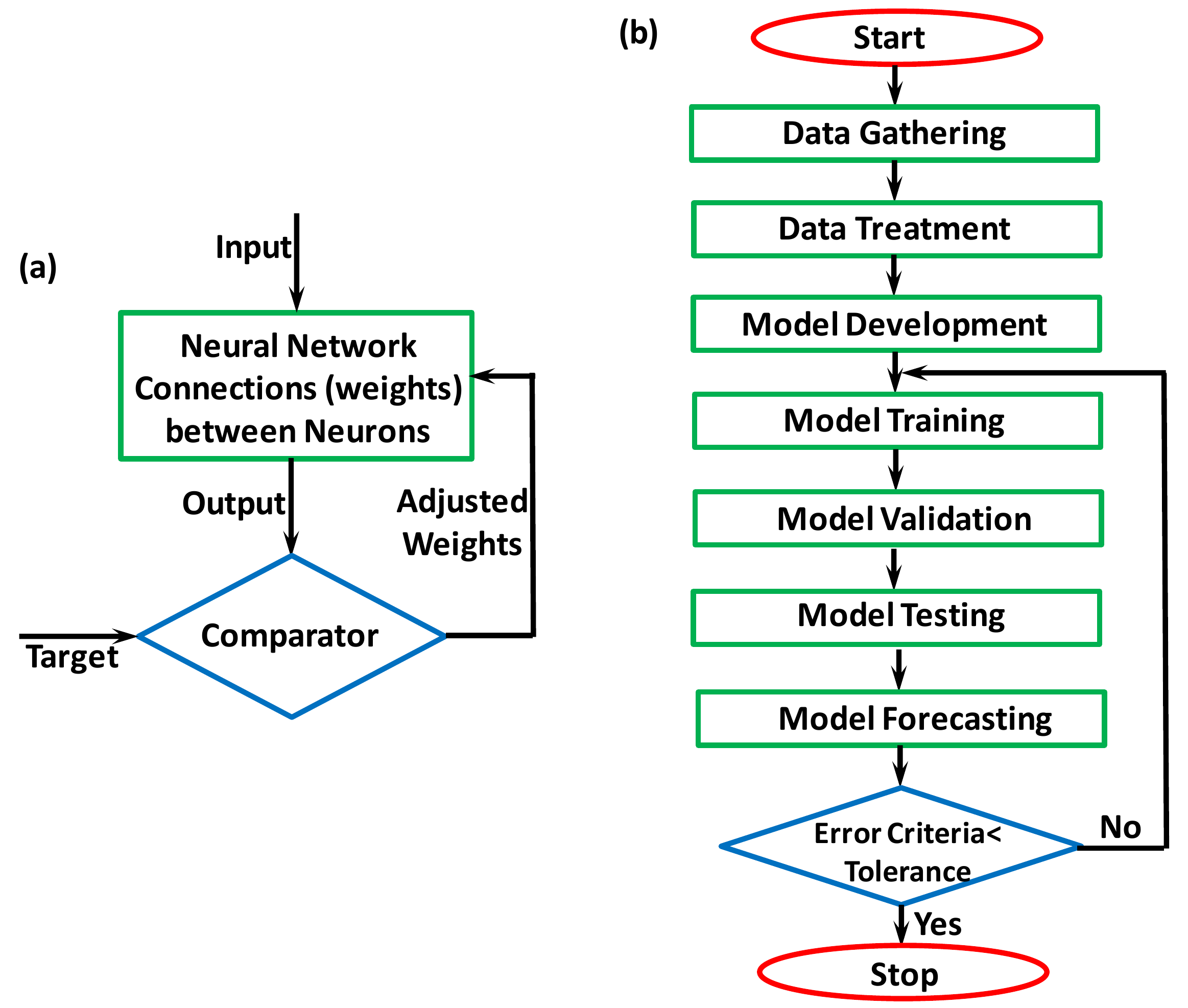} 
 \end{center}
\caption{(a) Neural Network Function (b) Flow Chart of the Model Process}
\label{fig:2}       
\end{figure*}

\begin{figure*}[h]
\begin{center}
 \includegraphics[scale=0.85]{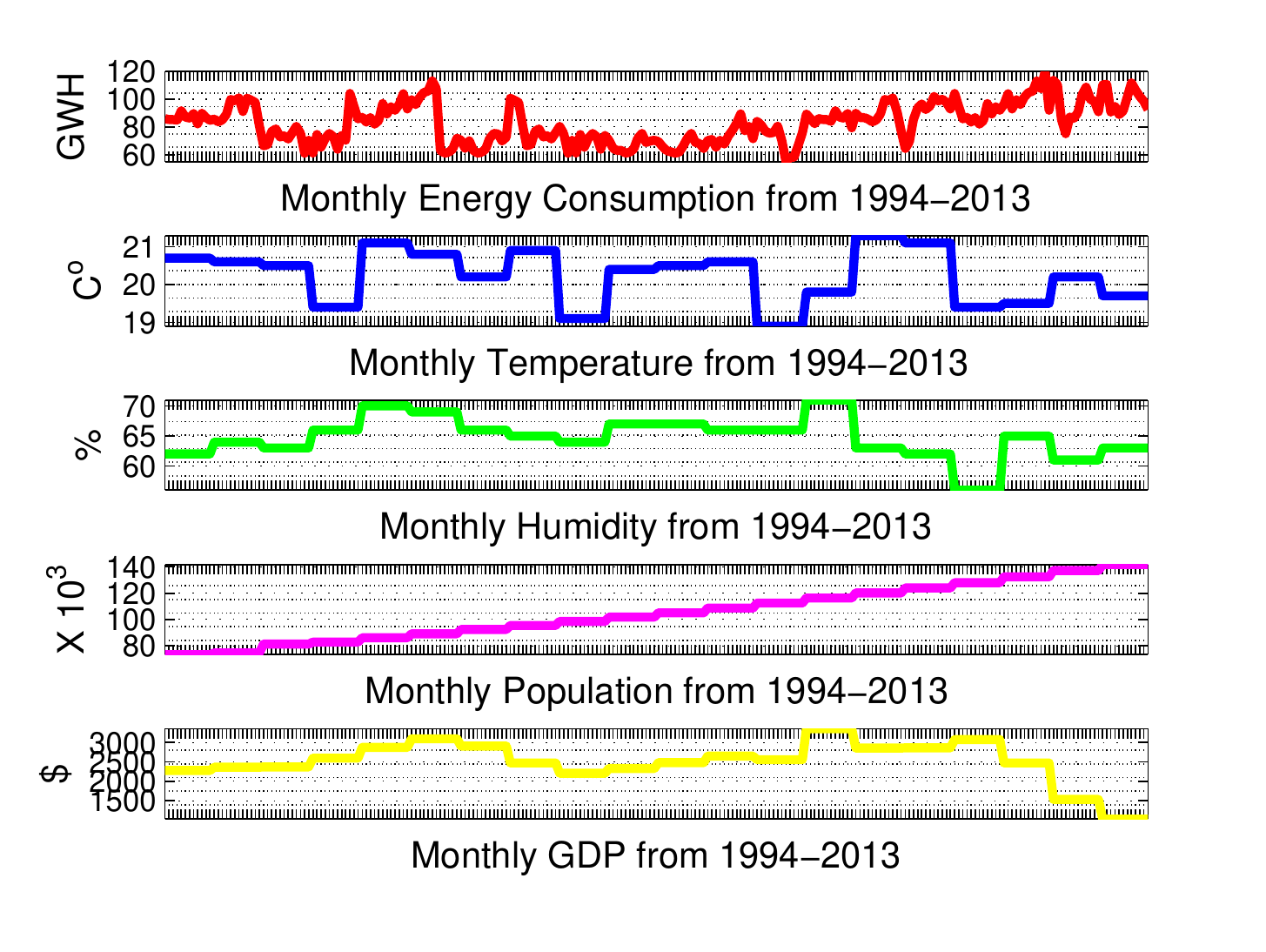}
 \caption{Historical Data for Gaza from 1994 - 2013}
\label{fig:3}       
\end{center}
\end{figure*}

In this section, we present the overall modeling process. The implementation of the ANN model can be described using the flow chart in Fig. \ref{fig:2}(b)\footnote{The error criteria used in this study are mean squared error (MSE), root mean squared error (RMSE), mean absolute error (MAE) and mean absolute percentage error (MAPE)}. Historical monthly data (historical energy consumption, mean atmospheric temperature, mean relative humidity, Population, GDP per Capita) has been gathered from years 1994 to 2013 from Gaza region as shown in Fig. \ref{fig:3}.
Based on the gathered data\footnote{Data was collected from the Palestinian Bureau of Statistic \cite{RefJ18} and the World bank website \cite{RefJ19}}, this study will develop a forecast model that predicts energy consumption for year 2012 and 2013. Using the ANN technique, training and learning procedures are fundamental in forecasting future events. The training of feed-forward networks is usually carried out in a supervised manner \cite{RefJ5}. With a set of data to be trained (usually extracted from the historical data), it is possible to derive an efficient forecast model. The proper selection of inputs for ANN training plays a vital role to the success of the training process. On the other hand, the learning process involves providing both input and output data, the network processes the input and compares the resulting output with desired result. The system then adjusts the weights which acts as a control for error minimization.
In order to minimize error, the process is repeated until a satisfactory criterion for convergence is attained. The knowledge acquired by the ANN via the learning process is tested by applying it to a new data set that has not been used before, called the testing set. It should now be possible that the network is able to make generalizations and provide accurate result for new data. Due to insufficient information, some networks do not converge. It is also noteworthy that over-training the ANN can seriously deteriorate forecasts. Also, if the ANN is fed with redundant or inaccurate information, it may destabilize the system.

Training and learning process should be thorough in order to achieve good results. To accurately forecast, it is imperative to consider all possible factors that influence electricity energy consumption, which is not feasible in reality. Electric energy consumption is influenced by a number of factors, which includes: historical energy consumption, mean atmospheric temperature, mean relative humidity, Population, GDP per Capita, PPP, etc. In this paper, different criteria were used to evaluate the accuracy of the ANN approach in forecasting electric energy consumption in Gaza. They include: mean squared error (MSE), root mean squared error (RMSE), mean absolute error (MAE) and mean absolute percentage error (MAPE).

A popular and important criterion used for performance analysis is the MSE. It is used to relay concepts of bias, precision and accuracy in statistical estimation. Here, the difference between the estimated and the actual value is used to get the error, the average of the square of the error gives an expression for MSE. The MSE criterion is expressed in Equation (\ref{eqn:1}).
\begin{equation}
\label{eqn:1}
MSE = \frac{1}{n}\sum_{i=1}^n (y_i -\widehat{y}_i)^2
\end{equation}
Where~$y_i$ is the actual data and~$\widehat{y}_i$ is the forecasted data. The RMSE is a quadratic scoring rule which measures the average magnitude of the error. The RMSE criterion is expressed in Equation (\ref{eqn:2}).
RMSE usually provides a relatively high weight to large errors due to the fact that averaging is carried out after errors are squared. This makes this criterion an important tool when large errors are specifically undesired.
\begin{equation}
\label{eqn:2}
RMSE = \frac{1}{n}\sqrt{\sum_{i=1}^n (y_i -\widehat{y}_i)^2}
\end{equation}
The MAE measures the average error function for forecasts and actual data with polarity elimination. Equation (\ref{eqn:3}) gives the expression for the MAE criterion used. The MAE is a linear score which implies that all the individual differences are weighted equally in the average.
\begin{equation}
\label{eqn:3}
MAE = \frac{1}{n}\sum_{i=1}^n |y_i -\widehat{y}_i|
\end{equation}
MAPE, on the other hand, measures the size of the error in percentage (\%) terms. It is calculated as the average of the unsigned percentage error. Equation (\ref{eqn:4}) gives the expression for the MAPE criterion used.
\begin{equation}
\label{eqn:4}
MAPE = \frac{1}{n}\sum_{i=1}^n \biggl|\frac{y_i -\widehat{y}_i}{y_i} \biggl| 100\%
\end{equation}
Validation techniques are employed to tackle fundamental problems in pattern recognition (model selection and performance estimation). In this study, 2-Fold and K-Fold cross validation techniques will be employed and the validation set will only be used as part of training and not part of the test set. The test set will be used to evaluate how well the learning algorithm works as a whole.

\section{Performance Evaluation}
\label{sec:4}
The forecast model was simulated to obtain results of the energy consumed for year 2012 and 2013 in Gaza. Table 1 compares the actual and the forecasted energy consumption for year 2012. 2-Fold and K-Fold cross validation techniques were used and the performance of the forecast model based on different error criteria is shown in Table 2. Similarly, Table 3 and 4 shows the results for year 2013. The results obtained have good accuracy and shows that the proposed ANN model can be used to predict future trends of electric energy consumption in Gaza.
\begin{table}[h]
\begin{center}
\begin{tabular}{|c  ||c|c|c|}
  \hline
  \textbf{Months} & \textbf{Actual - 2012} &	\textbf{Forecasted - 2012 (2 Fold)} &	\textbf{Forecasted - 2012 (k Fold)}\\ \hline \hline

  January & 113.5800 &	111.7538 &	119.9518  \\  \hline
  February & 110.5200 &	109.7213 &	109.1621 \\   \hline
  March & 85.3500	 &   86.6753	&    87.8927\\   \hline
  April & 74.9800	 &   75.2086	&    80.8937\\   \hline
  May & 87.1100 &    88.7946	&    90.4256\\   \hline
  June & 86.1000	 &   87.5484	&    88.8709\\   \hline
  July &  90.4600  &  90.0263	&    90.3327 \\   \hline
  August &  103.1600 &	102.9062	& 90.3061\\   \hline
  September  & 108.6600 &	107.4464	& 108.2806\\   \hline
  October  & 99.8500	&    100.7179	& 90.4634\\   \hline
  November  & 98.5000	&    99.9205	&    99.8268\\   \hline
  December  & 90.8400	&    93.4780	&    90.8787\\

  \hline
\end{tabular}
\end{center}
\caption{Forecasted Result for Year 2012}
\label{tab:1}

\begin{center}
\begin{tabular}{|c||c|c|c|c|}
  \hline
  \textbf{Year 2012} & \textbf{MSE} &	\textbf{RMSE} &	\textbf{MAE} &	\textbf{MAPE}\\ \hline \hline
2-Fold Cross Validation	& 1.21\%	& 1.10\%	& 1.21\%	& 120.91\% \\ \hline
K-Fold Cross Validation & 3.87\%	& 1.97\% &	3.87\%	& 386.54\% \\

  \hline

\end{tabular}
\caption{2-Fold and K-Fold Cross Validation for Year 2012}
\end{center}
\label{tab:2}
\end{table}

\begin{table}[h]
\begin{center}
\begin{tabular}{|c||c|c|c|}
  \hline
  \textbf{Months} & \textbf{Actual - 2013} &	\textbf{Forecasted - 2013 (2 Fold)} &	\textbf{Forecasted - 2013 (k Fold)}\\ \hline \hline

  January &	110.34	& 109.2616	& 110.0026	\\ \hline
  February &	110.67	& 109.3688	& 110.1573	\\ \hline
  March & 	90.776	& 92.9769 &	91.2542	\\ \hline
  April & 	94.66	& 97.1215 &	100.4115	\\ \hline
  May &	88.93 &	90.8877 &	89.2735	\\ \hline
  June &	91.443 &	93.723 &	94.6202	\\ \hline
  July &	101.82	& 102.7202 &	100.2111	\\ \hline
  August &	111.741	& 109.7194 &	110.7063	\\ \hline
  September  &	106.772	& 105.0375	& 108.7167	\\ \hline
  October  &	101.92	& 102.7778 &	100.2537	\\ \hline
  November  &	98.43	& 100.4408	& 100.0337	\\ \hline
  December  & 	92.53	& 90.4837 & 	96.5896	\\
  \hline
\end{tabular}
\end{center}
\caption{Forecasted Result for Year 2013}
\label{tab:3}

\begin{center}
\begin{tabular}{|c||c|c|c|c|}
  \hline
  \textbf{Year 2013} & \textbf{MSE} &	\textbf{RMSE} &	\textbf{MAE} &	\textbf{MAPE}\\ \hline \hline
2-Fold Cross Validation	& 1.74\%	& 1.32\%	& 1.74\%	& 173.76\% \\ \hline
K-Fold Cross Validation & 1.88\%	& 1.37\% &	1.88\%	& 187.65\% \\
 \hline

\end{tabular}
\caption{2-Fold and K-Fold Cross Validation for Year 2013}
\end{center}
\label{tab:4}
\end{table}

\begin{figure*}[Ht]
\begin{center}
 \includegraphics[scale=0.8]{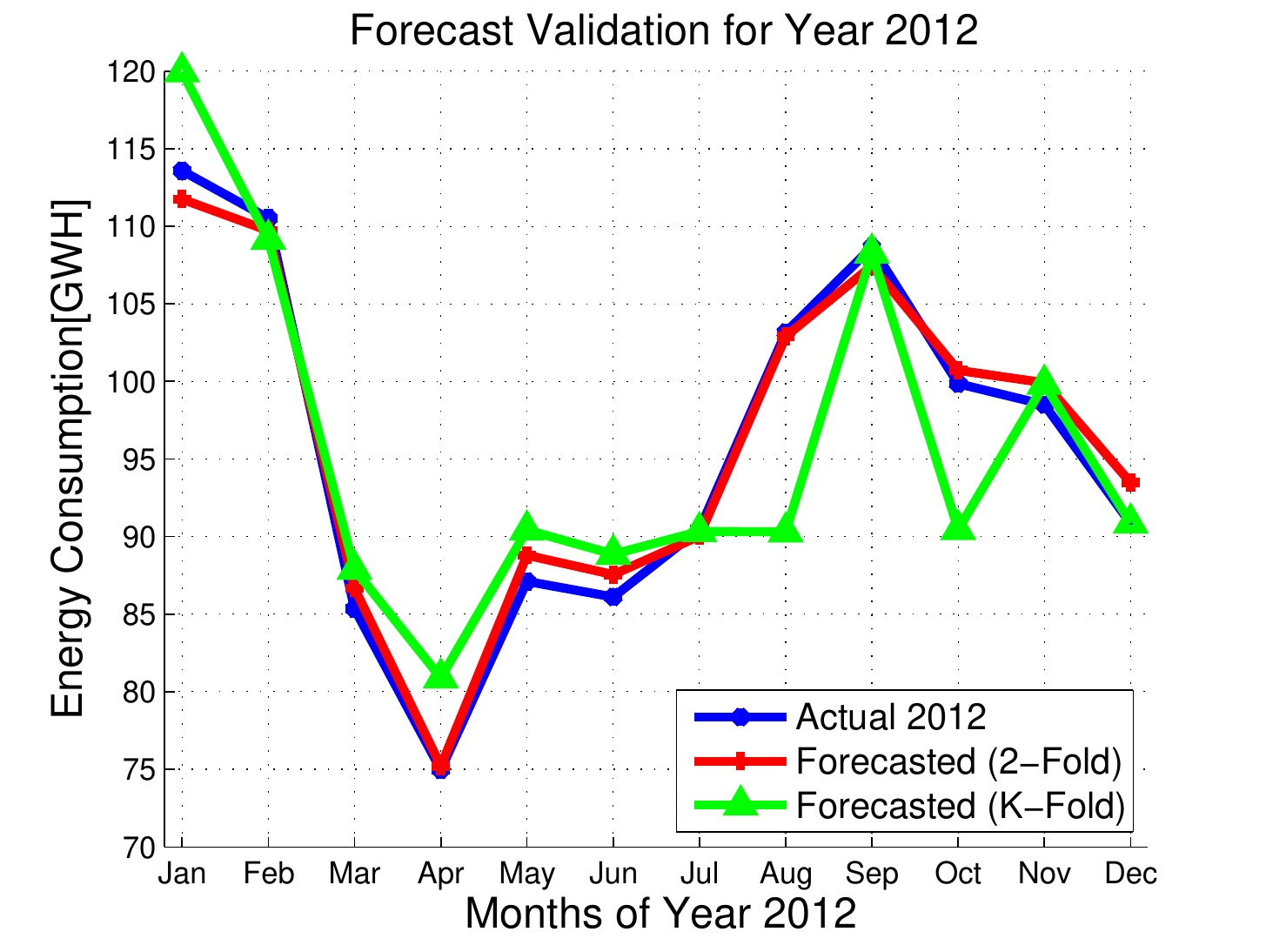}
 \caption{Plot of Forecast Validation for Year 2012}
\label{fig:4x}       
\end{center}
\end{figure*}

\begin{figure*}
\begin{center}
 \includegraphics[scale=0.8]{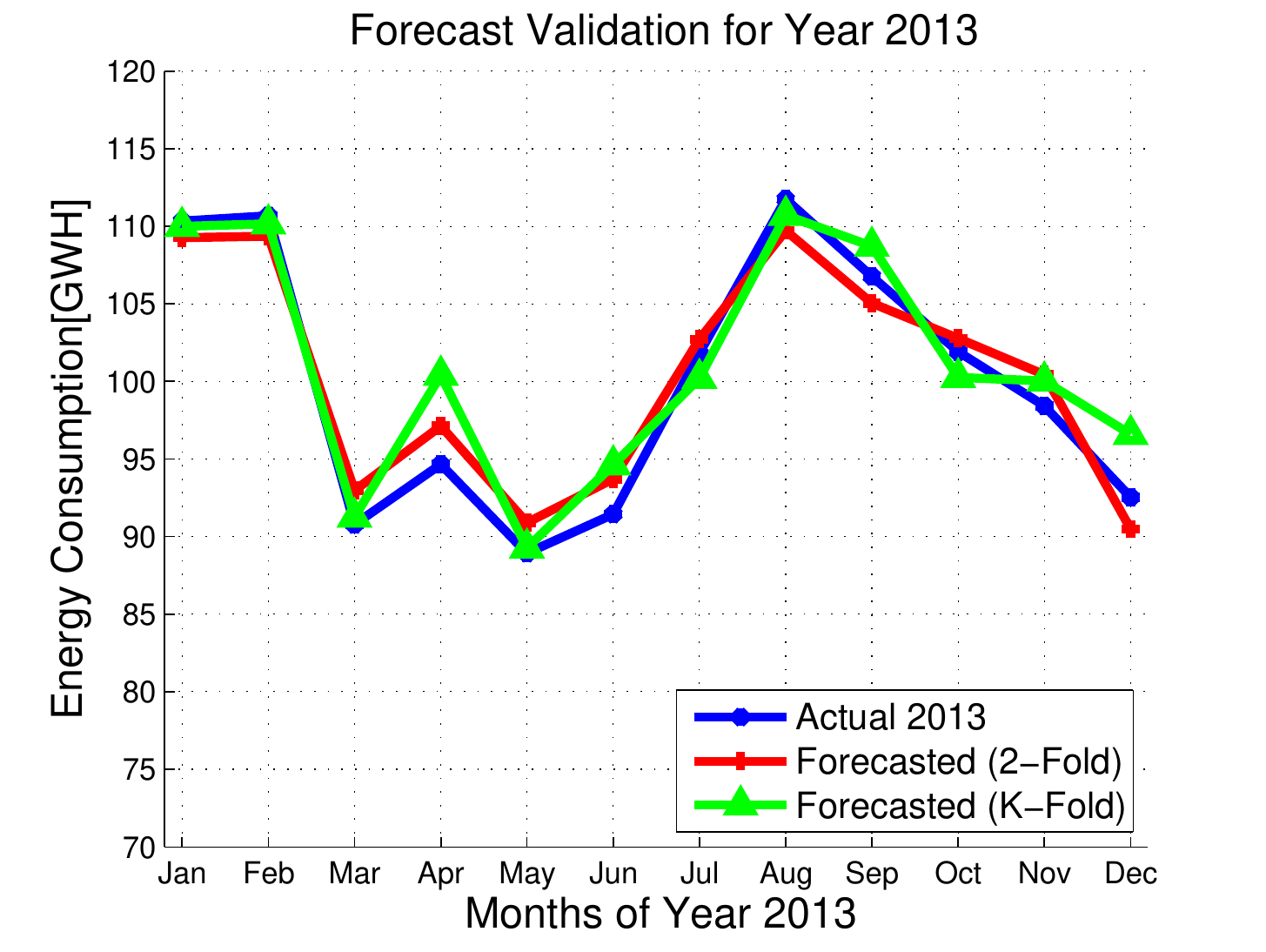}
 \caption{Plot of Forecast Validation for Year 2013}
\label{fig:4y}       
\end{center}
\end{figure*}
\section{Conclusion}
\label{sec:5}
In this paper, an ANN model to forecast electric energy consumption in the Gaza strip was presented. To the best of our knowledge, this is the first of it's kind in existing literature. Based on the performance evaluation, the error criteria were within tolerable bounds. Empirical results presented in this paper indicates the relevance of the proposed ANN approach in forecasting electric energy consumption. Future works will consider other forecasting techniques.




\end{document}